%% file: arxiv.tex
\documentclass{article}
\usepackage{graphicx}

\usepackage[preprint]{neurips_2023}

\usepackage[utf8]{inputenc} % allow utf-8 input
\usepackage[T1]{fontenc}    % use 8-bit T1 fonts
\usepackage{hyperref}       % hyperlinks
\usepackage{url}            % simple URL typesetting
\usepackage{booktabs}       % professional-quality tables
\usepackage{amsfonts}       % blackboard math symbols
\usepackage{nicefrac}       % compact symbols for 1/2, etc.
\usepackage{microtype}      % microtypography
\usepackage{xcolor}         % colors
\usepackage{color}
\usepackage{wrapfig} % wrapped table
\usepackage{boxedminipage}
\usepackage{framed}
\usepackage{soul}
\usepackage{xspace}
\usepackage{booktabs} % professional-quality tables
\usepackage{courier} % light font weight
\usepackage{makecell}
\usepackage{xcolor}
\usepackage{multirow} 
\usepackage{graphicx}
\usepackage[export]{adjustbox}
\usepackage{float}
\usepackage{epsfig}
\usepackage{graphicx}
\usepackage{amsmath}
\usepackage{amssymb} % http://ctan.org/pkg/amssymb
\usepackage{caption}
\usepackage{xparse} % data cards
\usepackage{wrapfig} % wrapped table
\usepackage{tabularx}
\usepackage{subcaption}
\usepackage{listings}
\usepackage{natbib}
\setcitestyle{numbers,square}
\definecolor{my_green}{RGB}{51,102,0}
\definecolor{my_yellow}{RGB}{255,165,0}
\definecolor{my_red}{RGB}{204, 0, 0}

\newcommand{\red}[1]{\textcolor{red}{#1}}

\newcommand{\yellow}[1]{\textcolor{my_yellow}{#1}}
\newcommand{\blue}[1]{\textcolor{blue}{#1}}

\newcommand{\pan}[1]{\textcolor{cyan}{#1}}

\newcommand{\CP}{\blue{\textbf{Context Prompt:}}}
\newcommand{\UP}{\pan{\textbf{User Prompt:}}}
\newcommand{\GP}{\red{\textbf{Generated Response:}}}
\newcommand{\LP}{\red{\textbf{LLaVA Response:}}}

\definecolor{commentcolor}{RGB}{110,154,155}
\definecolor{inputcolor}{RGB}{102, 59, 12}
\definecolor{problemcolor}{RGB}{242,242,242}
\definecolor{resultcolor}{RGB}{252,228,215}
\definecolor{contextcolor}{RGB}{160,198,157} % add a "#" before the input text "#1"

\newcommand{\Context}[1]{\colorbox{contextcolor}{\parbox{\linewidth}{#1}}
}

\hypersetup{
    colorlinks=true,
    urlcolor=magenta,
}
\usepackage{caption}
\usepackage{subcaption}
\usepackage{multirow}
\usepackage{xcolor}
\usepackage{tcolorbox}
\usepackage{mathptmx}
\usepackage{wrapfig}
\newcommand{\modelname}{LI3D}
\tcbuselibrary{breakable}
\newcommand{\eg}{\textit{e}.\textit{g}.}

\newcommand{\ie}{\textit{i}.\textit{e}.}

\title{Towards Language-guided Interactive 3D Generation: LLMs as Layout Interpreter with Generative Feedback}

\author{%
Yiqi Lin$^{1}$ 
\quad Hao Wu$^{1}$ 
\quad Ruichen Wang$^{2}$ 
\quad Haonan Lu$^{2}$ 
\quad Xiaodong Lin$^{3}$
\\
\quad \textbf{Hui Xiong}$^{1,4}$
\quad \textbf{Lin Wang}$^{1,4}$\footnotemark[1]
\\
$^{1}$AI Thrust, HKUST(Guangzhou)
\quad $^{2}$OPPO
\quad $^{3}$Rutgers University
\quad $^{4}$ Dept. of CSE, HKUST
\\
{\tt\small ylin933@connect.hkust-gz.edu.cn
\quad hwubx@connect.ust.hk}
\\
{\tt\small\{wangruichen, luhaonan\}@oppo.com
\quad lin@business.rutgers.edu}
\\
\tt\small{\{xionghui, linwang\}@ust.hk}
}

\begin{document}

\maketitle
\footnotetext[1]{Corresponding author.}

\begin{abstract}
Generating and editing a 3D scene guided by natural language poses a challenge, primarily due to the complexity of specifying the positional relations and volumetric changes within the 3D space. Recent advancements in Large Language Models (LLMs) have demonstrated impressive reasoning, conversational, and zero-shot generation abilities across various domains. Surprisingly, these models also show great potential in realizing and interpreting the 3D space. In light of this, we propose a novel language-guided interactive 3D generation system, dubbed \textbf{\modelname}, that integrates LLMs as a 3D layout interpreter into the off-the-shelf layout-to-3D generative models, allowing users to flexibly and interactively generate visual content. Specifically, we design a versatile layout structure base on the bounding boxes and semantics to prompt the LLMs to model the spatial generation and reasoning from language. Our system also incorporates LLaVA, a large language and vision assistant, to provide generative feedback from the visual aspect for improving the visual quality of generated content. We validate the effectiveness of \modelname, primarily in 3D generation and editing through multi-round interactions, which can be flexibly extended to 2D generation and editing. Various experiments demonstrate the potential benefits of incorporating LLMs in generative AI for applications,~\eg, metaverse. Moreover, we benchmark the layout reasoning performance of LLMs with neural visual artist tasks, revealing their emergent ability in the spatial layout domain.

\end{abstract}

\vspace{-10pt}
\section{Introduction}
\vspace{-10pt}
There has been a growing amount of work~\cite{poole2022dreamfusion,metzer2022latent,lin2022magic3d,wang2022score} using powerful text-to-image pre-trained models~\cite{radford2021learning,saharia2022photorealistic} to generate high-quality 3D scenes based on the text inputs. 
Despite the impressive results reported in the literature, it still remains a challenge of achieving a level of grace and flexibility in generating and manipulating scenes through open-vocabulary natural language that is on par with human interaction. 
To enhance human-computer interaction in image generation, researchers~\cite{zhou2022tigan,xia2021tedigan,cui2023i2edit} have attempted to utilize pre-trained cross-modal models, \eg, CLIP~\cite{radford2021learning} or generative models, \eg, StyleGAN~\cite{karras2019style} to generate images through continual linguistic instructions.
However, for real-world applications, such models suffer from limited capacities to comprehend and reason the natural language input in the open-vocabulary domain.
Moreover, these works have solely focused on the 2D generation and do not account for the challenges of rendering 3D scenes, particularly related to the arrangement and editing of multiple 3D objects.

Meanwhile, recent advances in Large language models (LLMs), such as PaLM~\cite{chowdhery2022palm}, LLaMA~\cite{touvron2023llama} and GPT-4~\cite{openai2023gpt}, exhibit exceptional abilities in carrying out natural language reasoning tasks and demonstrate impressive conversational, reasoning, and zero-shot generation abilities across diverse domains.
More remarkably, LLMs have shown eminent potential in realizing and interpreting the 3D space through codes~\cite{bubeck2023sparks}, thus making it possible to be a strong linkage between the natural language and the 3D generation modeling without relying on 3D datasets.

However, it is non-trivial to directly apply LLMs to solve such complex interactive 3D generation tasks without considering the proper cooperation with off-the-shelf generative models.
In light of this, we propose to use the layout as the interface to bridge the LLMs and generative models, \ie, the LLMs interpret natural language into a 3D layout across multi-round interactions and the 3d generative models, \eg, CompoNeRF~\cite{lin2023componerf} use the 3D layout as conditional input to perform layout-to-3D generation.
To achieve this, we design a language-guided interactive 3D generation system, dubbed \textbf{\modelname}, that defines a versatile layout structure based on bounding boxes and semantics for LLMs to interpret.
This allows our system to not only generate 3D scenes from the stretch but also enable the editing of various objects within the generated scene. 
Furthermore, we integrate the LLMs~\cite{liu2023visual,gong2023multimodal,zhu2023minigpt} boosted by visual instruction tuning, \eg, LLaVA~\cite{liu2023visual} in our case, into our system to improve the stability of the generated 3D scenes.
Because recent works~\cite{metzer2022latent, lin2023componerf,cheng2022sdfusion} have demonstrated the instability of generating complex 3D scenes when relying solely on general descriptions. 
This instability often results in unsatisfactory visual quality and potential confusion in modeling multiple objects.
Firstly, the LLaVA~\cite{liu2023visual} acts as the verifier to predict the visual similarity between generated rendered image and the description.
Secondly, for the low-quality generated content, the LLaVA provides a more detailed description of the scene based on the image for the LLMs interpreter to refine the layout.

Our extensive empirical evaluations showcase the versatility and efficacy of \modelname~in both 2D and 3D generation tasks. 
We evaluate the \modelname~using the i-CLEVR benchmark~\cite{el2019tell} to assess its robustness in generating layouts based on sequential language instructions. 
The results indicate that \modelname~can achieve a high recall rate in recognizing objects and a strong reasoning capacity of 3D spatial relationships modeling under multi-round interactions. 
The visual results show \modelname~can generate 3D scenes that are consistent with the user's natural language input in multiple rounds. We also conduct studies on 3D single object generation, which reveal the~\modelname~having knowledge about the components of individual 3D objects and being able to understand and edit fine-grained parts following user language input.
Additionally, we extend our pipeline to 2D image generation and demonstrate promising results in multi-round 2D image generation and editing.

In summary, our proposed interactive 3D generation system, \textbf{\modelname}, utilizes the capabilities of LLMs to interpret natural language input and generate 3D layouts, allowing for highly customized and realistic 3D scenes. Our evaluation results demonstrate the robustness and versatility of \modelname~in both 2D and 3D interactive generation tasks based on natural language input and multi-round interactions.

\begin{figure}[t!]
    \centering
    \includegraphics[width=\linewidth]{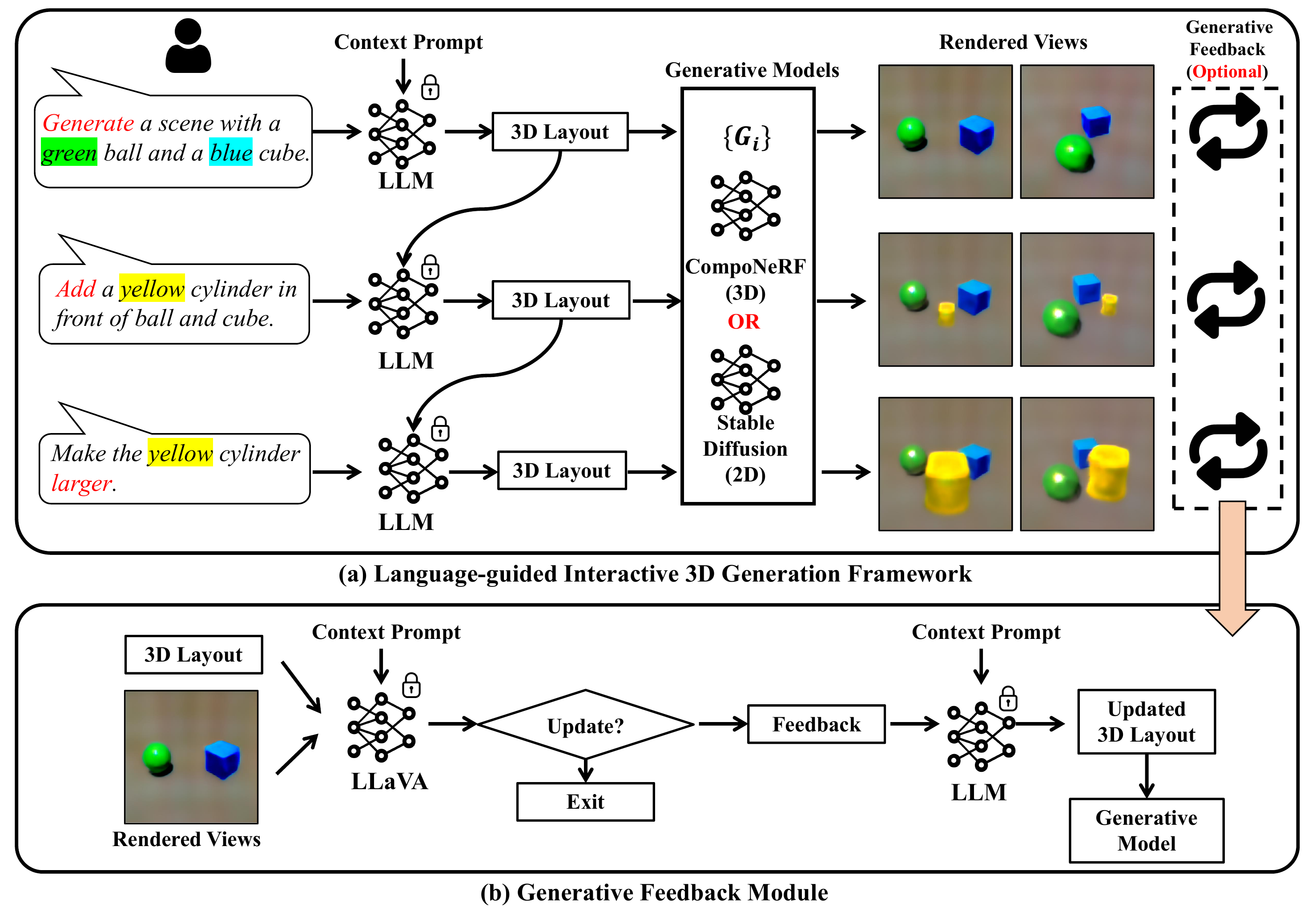} 
    \caption{The overall system of \modelname.
    (a) \modelname~utilize the LLM to interpret language input into the 3D layout for layout-to-3D generative model (CompoNeRF~\cite{lin2023componerf}) as conditional input (Sec.~\ref{sec:li3d}), which can also be extended to the image domain by several adaptions (Sec.~\ref{sec:2d}). (b) The LLaVA~\cite{liu2023visual} can be integrated into \modelname~to predict generation quality and provide detailed description feedback for LLM to update the layout that fails to generate satisfactory content (Sec.~\ref{sec:fb}).
    }
    \label{fig:framework}
\end{figure}

\vspace{-5pt}
\section{Proposed Method}
\noindent \textbf{Overview of our \modelname:}
An overview of our proposed system is depicted in Fig.~\ref{fig:framework}.
The key idea is to utilize the spatial domain (2D or 3D) knowledge, the remarkable conversational competency, and reasoning capabilities within recent powerful LLMs~\cite{ouyang2022training,bubeck2023sparks,thoppilan2022lamda} to interpret the user language into the 3D layout as input to the conditional generative models.
Language-guided interactive 3D generation aims to generate 3D content following multi-round user language inputs, which requires the model to understand the relationship within the input language and be capable of \textit{interactively} generating 3D content.
In our work, we bridge the gap between LLMs and the 3D generative models by considering the generated 3D content as a composition of multiple objects or parts that can be formalized by coarse 2D or 3D boxes with semantic descriptions in a scene.
Therefore, \modelname~leverages the LLMs to interpret the user language into the 3D layout for guiding the off-the-shelf generative models to perform conditional 3D generation.
Using the 3D layout to represent a 3D scene makes the LLMs maintain the context of the current conversation more precisely and reason the language input in a more standardized way.
Moreover, we incorporate the LLaVA~\cite{liu2023visual} to provide feedback and generate more details descriptions based on the rendered image for LLMs, allowing for updating the interpreted layout when it fails to generate satisfactory results.
Empirically, the conditional layout with more details can further improve the visual quality of generated content.

\subsection{Language-guided Interactive 3D Generation Framework}
\label{sec:li3d}
As shown in Fig.~\ref{fig:framework}(a), \modelname~employs a natural language interpreter to understand the user input and generate a layout that includes concept semantics, \eg, blue cube, and location information, \eg, the center of the scene.
Each element in the layout is a coarse box with a semantic description of a scene.
This layout is then used as the conditional input to generative modules to produce visual content.
Formally, given a pre-defined context prompt $p$, input language $x$, and generative module $G$, the natural language interpreter $I$ generates a corresponding layout $l$ by responding to the user's query.
In our work, the interpreter $I$ is the LLM, which is prompted to generate a layout given a 2D or 3D generation task setup (\textit{See details in Sec.~\ref{sec:setup} and Appendix}).
The generative module $G$ consists of a set of pre-built modules: $\{G_i\}$, each corresponding to a type of models, \eg,  Stable Diffusion~\cite{rombach2022high} or CompoNeRF~\cite{lin2023componerf}.
Eq.~\ref{eq:1} provides a formal definition of the layout $l$:
\begin{equation}
    l = I(x_0;p)
    \label{eq:1}
\end{equation}
where $x_0$ is the initial language input for 3D generation from scratch.
Given the generated layout, the corresponding generative modules for each step are then executed to obtain generated results $G(l)$.
To improve the consistency across multi-round interactive generation, we first detect the layout modification compared to the previous step and then apply the pre-defined rules to update the generated result using generative models.
For time step $t$, the rendered views $r_t$ is generated by:
\begin{equation}
r_t = G(l_{t-1}, I(x_t;p;l_{t-1}))
\end{equation}

During an interaction session, we use the rule-based operation to perform interactive editing based on the principle that uses semantic description as the unique identifier for each 3D box.
When detecting the addition of a new box (i.e., instance) to the layout, we only initialize a new instance based on the previously generated results, while keeping the rest unchanged.
When detecting edits to the attribute in the semantic description, we further fine-tune the modified object to align with the edited description.
When detecting edits to the location, we can simply modify the transformation matrix associated with the 3D location.
For interactive image generation, we introduce both Stable Diffusion~\cite{rombach2022high} and Segment Anything models~\cite{kirillov2023segment} to generate, extract and manipulate the object based on the layout. More details are illustrated in Sec.~\ref{sec:2d}.

\subsection{Generative Feedback Module}
\label{sec:fb}
Recent research~\cite{metzer2022latent,lin2023componerf} has shown that generating complex 3D scenes can be unstable when using less informative descriptions, leading to unsatisfactory visual quality and confusing multi-object modeling results.
Based on the empirical study, we find that \modelname~tends to generate the layout with a general description, which may trigger the aforementioned issues.
To this end, we propose a generative feedback module to automatically assess the suitability of the generated 3D scene and ensure proper 3D placement using LLaVA~\cite{liu2023visual} instead of responses from the user. 
It employs the LLaVA as the verifier $V$ to explicitly predict whether the rendered views satisfy the description provided in the layout using the context prompt $\hat{p}$, as shown in Fig.~\ref{fig:framework} (b).
Then, the feedback $b_{t} = V(l, p,{r}_t)$ based on the rendered view ${r}_t$ is utilized to help LLMs update the layout with more details when these views fail to match the description through LLaVA.
Formally, the updated rendered view ${r}_t^{*}$ is calculated by: 
\begin{equation}
    {r}_t^{*}= G(l_{t-1}, I(x_{t};p;b_{t}))
\end{equation}
The generative feedback module can be seamlessly integrated at every step of the~\modelname~as an enhancer. 

\noindent \textbf{Discussion:}
Recent research~\cite{bubeck2023sparks} has showcased that the LLMs, \eg,~\cite{chowdhery2022palm,openai2023gpt}, can synthesize information from different domains or modalities and can apply knowledge across different contexts.
Specifically, the LLMs are able to generate primitive graphics or geometries through codes.
An intuitive idea is to take such primitives as the conditional input for the generative models.
However, it is nontrivial to compose the off-the-shelf generative models with the code interface in an interactive way.
Therefore, it motivates us to propose \modelname~with a simple yet effective intermediate representation design to interpret the user intention from language through LLMs.
Using layouts with boxes to model scenes, \modelname~facilitates precise element-level decomposition and editing, allowing for more accurate context preservation in multi-round scenarios compared to code formats or plain language.

\section{Empirical Evaluation and Analysis}
As there lack of benchmark studies for the interactive 3D generation task, we conduct a quantitative evaluation and case study to demonstrate the versatile use of LLMs as the layout interpreter by answering the following questions:
\textbf{1}) How well do LLMs understand the language instruction in the spatial domain?
\textbf{2)} How well do LLMs learn about the common scene about 3D scene understanding or single object?
To address these questions, we first introduce a benchmark with the dataset from the interactive visual generation to evaluate the potential of layout from language instruction.
Then, we provide the case study results, analysis and failure cases that could lead to future improvement.

\subsection{Experiment Setup}
\label{sec:setup}
\noindent \textbf{Implementation Details}
We implement the LLMs with ChatGPT~\cite{ouyang2022training} (OpenAI "gpt-3.5-turbo" version) and the LLaVA-13B~\cite{touvron2023llama} as our generative feedback module.
Once a 3D layout response is generated, we feed it into the CompoNeRF~\cite{lin2023componerf} for generative rendering on a single RTX3090. For more details of CompoNeRF please refer to the paper~\cite{lin2023componerf}.
For the generation stage, \ie, the initial user input, we train the CompoNeRF from scratch with 8000 iterations.
For the following editing stages, we use a rule-based strategy to update previously generated 3D content based on CompoNeRF.
During a user interaction session, we use the object description as their unique id for each local NeRF.
When we detect a new instance is added to the layout, we first train the single local NeRF in 3000 iterations and then joint fine-tunes the whole scene with 6000 iterations.
When the layout contains duplicated object description, we use the minimal location distance for the matching modified object from the previous generated results to perform location editing.

\noindent \textbf{Prompt Design}
Similar to VisualChatGPT~\cite{wu2023visual}, we design a context template for prompting the layout generation.
Our case study focuses on two scenarios: scene generation and object generation. To optimize generation quality in each scenario, we developed two types of context prompts with different emphases, as shown in Tab.~\ref{prompt}.
For the feedback generation, we randomly sample a rendered image and the context prompt with layout information, \ie description of scene or object, as input to predict the matching similarity between the image and description.
When the LLaVA generates a negative response, we further use LLaVA to generate a more detailed description of the current content as feedback input for LLMs to update the layout. 
More details are provided in Appendix.

\subsection{Robustness of Layout Generation from Language}
To evaluate layout generation following the language input qualitatively, we utilize a subset of i-CLEVR~\cite{el2019tell}, a dataset designed for the Generative Neural Visual Artist (GeNeVA) task, as the benchmark.
Each example in i-CLEVR~\cite{el2019tell} dataset consists of a sequence of 5 (image, instruction) pairs. Starting from an empty background, each language instruction describes an object to add to the canvas in terms of its shape and color. 
The instruction also describes where the object should be placed relative to existing objects in the scene. 
We sample 50 sequential instructions and their ground-truth arrangement to evaluate the robustness of LLMs following instructions.
LLM is asked to generate the scene layout based on the textual instruction with multi-round sequential instruction (max to 5).
More details about the evaluation can be found in the appendix.

To compute a similarity metric between the generated layout and the ground truth, we use recall and relational similarity (rsim) proposed in~\cite{el2019tell} as the evaluation metric.
The recall is the objects recalled in the generated layout based on the ground truth.
The relational similarity metric quantifies the overlap between the ground truth relations and the generated layout by counting the number of shared relations.
The relations are defined by the left-right and front-back relations between each object.
Tab.~\ref{tab:num} reports the mean of the different time-step values for each example over the sampled sub-dataset.
It shows that the LLMs can recognize the object identities from multi-round language input with high accuracy and achieve a strong reasoning capacity in the 3D spatial domain for object arrangement.
This phenomenon motivates us to further develop LLMs as versatile layout interpreters for 3D and 2D interactive generation tasks.

\input{prompt_design}

\begin{table}[t]
\vspace{-8pt}
\centering
\refstepcounter{table}
\label{tab:num}
\caption{Quantitative results on the i-CLEVR datasets.}
\begin{tabular}{l|ccccc} 
\hline
 & Round-1 & Round-2 & Round-3 & Round-4 & Round-5 \\ 
\hline
Recall & 98.0 & 98.0 & 98.0 & 97.9 & 97.9 \\
rsim($E_{G_{gt}}$,$E_{G_{gen}}$) & 97.9 & 84.1 & 78.2 & 71.0 & 66.0 \\
\hline
\end{tabular}

\end{table}

\subsection{Qualitative Results of Interactive 3D Generation}

\begin{figure}[!t]
    \centering
    \includegraphics[width=0.95\linewidth]{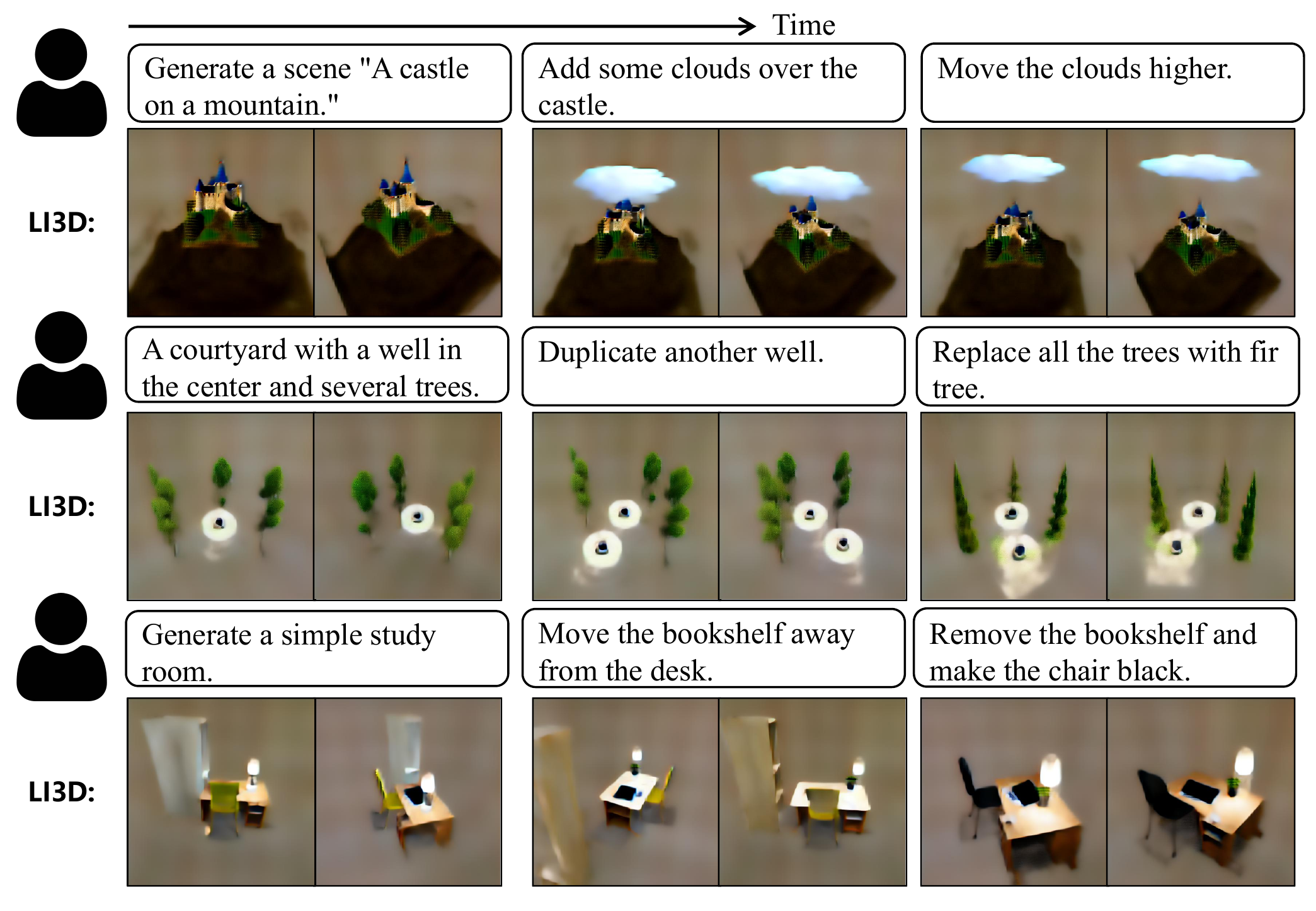}
    \caption{Multiple rounds of interaction for 3D scene generation between users and \modelname.}
    \label{fig:scene}
\end{figure}

\noindent \textbf{3D Scene Generation} Results are shown in Fig.~\ref{fig:scene}. 
In this case, the user can generate a complex compositional scene with a simple prompt.
Our \modelname~can generate reasonable spatial layouts as responses.
The generated scene involves different scales from a mountain to a small room.
Moreover, the editing instruction can include adding/removing objects and adjusting/duplicating objects.
However, in some instances, the LLMs fail to generate a reasonable layout without the user adjustment, \eg, the book self in the third case.
Nevertheless, \modelname~shows the great potential of using LLMs as the interface for the interactive 3D generation task based on user language.

\begin{figure}[t]
    \centering
    \includegraphics[width=0.95\linewidth]{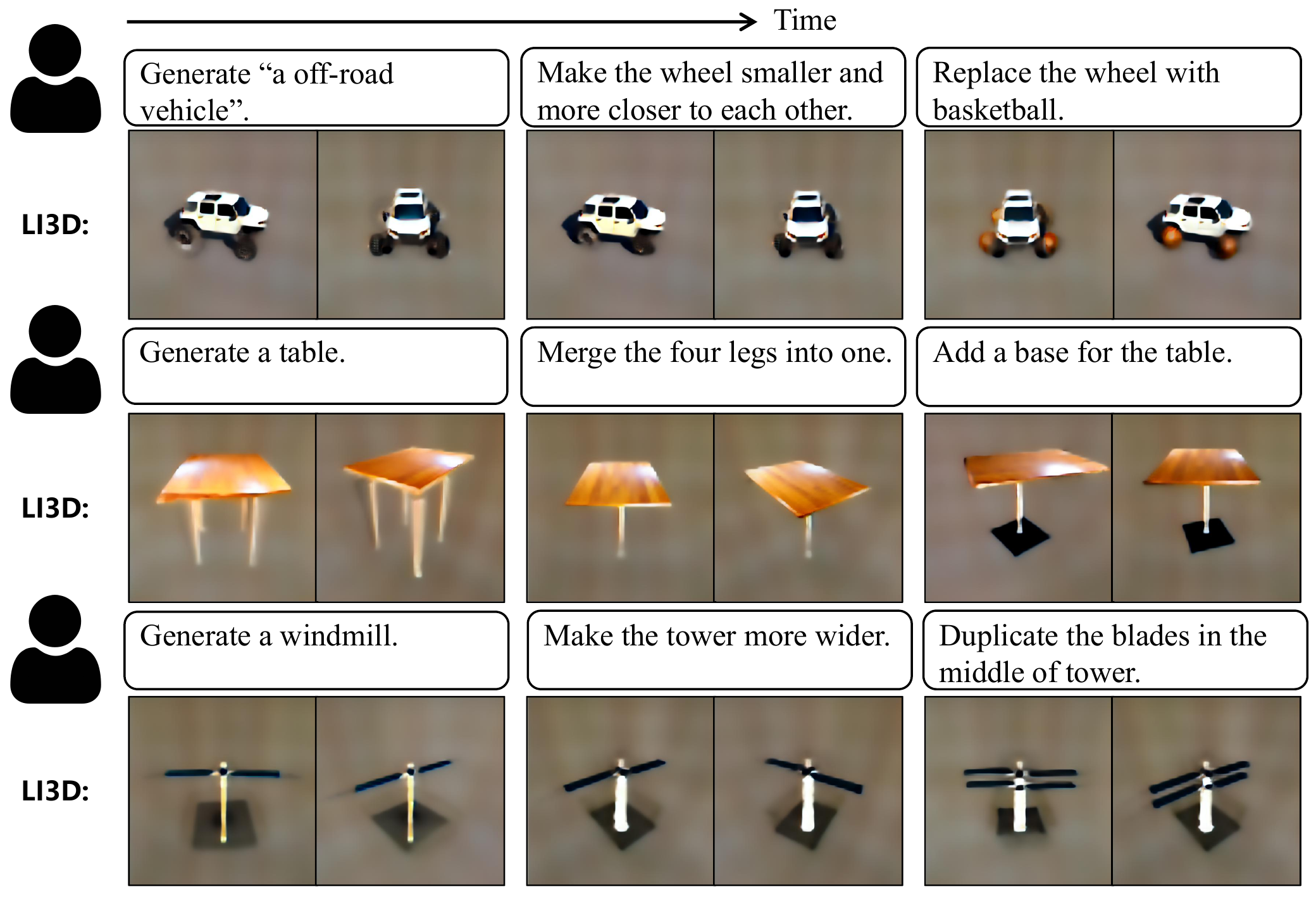}
    \caption{Multiple rounds of interaction for 3D single object generation between users and \modelname.}
    \label{fig:obj}
\end{figure}

\noindent \textbf{Single Object Generation} As shown in Fig.~\ref{fig:obj},~\modelname~also enables users to generate individual objects from simple compositional parts using language. The results of our \modelname~reveal that LLMs, which are trained solely on text data, are capable of learning about the concept of object decomposition.
This suggests that LLMs have the potential as knowledge engines in future work for 3D generation modeling.
Based on the decomposed parts in the layout, our \modelname~allows editing operations including adding, removing, and adjusting object parts.
Despite the impressive results~\modelname~achieved, \modelname~still can not decompose the object following the physical rules even with simple structure, \eg, the wheels are out of the vehicle in certain views.

\begin{figure}[t]
    \centering
    \includegraphics[width=0.95\linewidth]{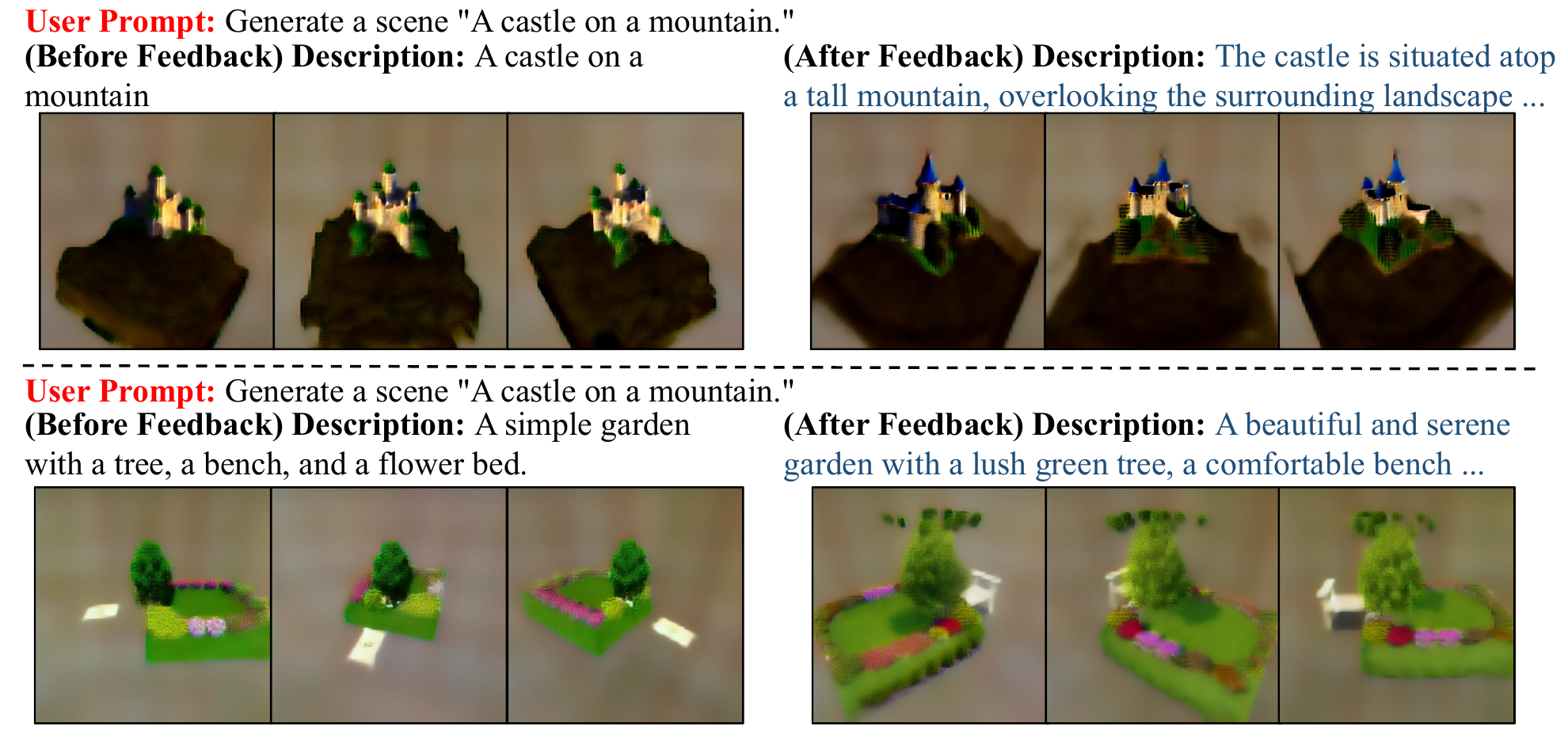}
    \caption{Ablation study of generative feedback module.}
    \label{fig:ab}
\end{figure}

\noindent \textbf{Ablation Study of Generative Feedback}
To verify the effectiveness of our generative feedback module, we perform the ablation study in Fig.~\ref{fig:ab}.
Specifically, we provide two cases before and after using the generative feedback module.
The results suggest that the generated general layout may not be accurately captured in generative models resulting in low visual quality.
However, when more detailed description feedback from LLaVA was incorporated, the generated layout contained richer descriptions of the scene leading to more satisfactory generated content.

\begin{figure}[t]
    \centering
    \includegraphics[width=\linewidth]{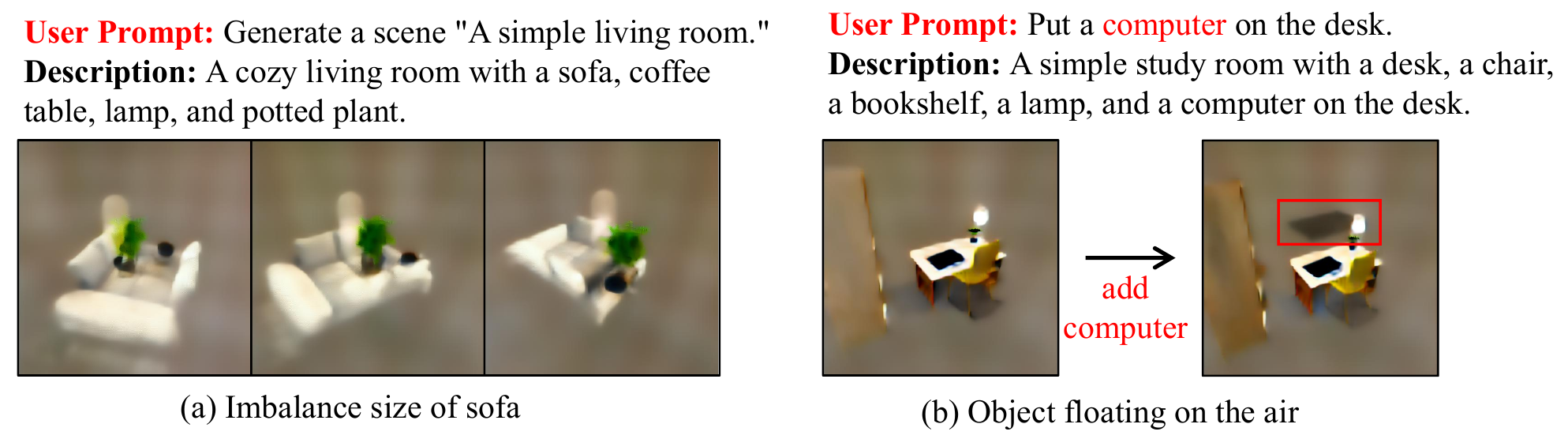}
    \caption{Failure cases of \modelname.}
    \label{fig:fail}
\end{figure}

\noindent \textbf{Failure Cases \& Limitation}
While \modelname~represents a significant advancement towards language-guided interactive 3D generation using the LLMs, it still has limitations that need to be addressed.
As shown in Fig.~\ref{fig:fail}, the LLMs may generate a scene with imbalance size among different objects contrary to common sense and reasoning of the layout considering the physical rules in the real world.
For these cases, our generative feedback module also fails to provide correct feedback due to their generation quality.
These failure cases highlight the importance of providing constraints based on common sense in more realistic and precise 3D content generation.
Besides, the nondeterministic behavior of the LLMs may result in different generation contents during different trials or the non-consistent or mistakenly generated layout across a long sequence of interactions.
Moreover, the foundational 3D generative models, such as CompoNeRF~\cite{lin2023componerf}, are still unstable in constructing certain open-vocabulary single objects, especially irregular objects, and can not achieve real-time interaction performance limited by their computation cost. 
Although our interactive system can generate and edit complex scenes based on language, the low resolution of generated scenes might have limited performance when involving a large number of objects or objects having extreme size differences.

\subsection{Extension on Interactive Image Generation}
\label{sec:2d}
\begin{figure}[t]
    \centering
    \includegraphics[width=0.95\linewidth]{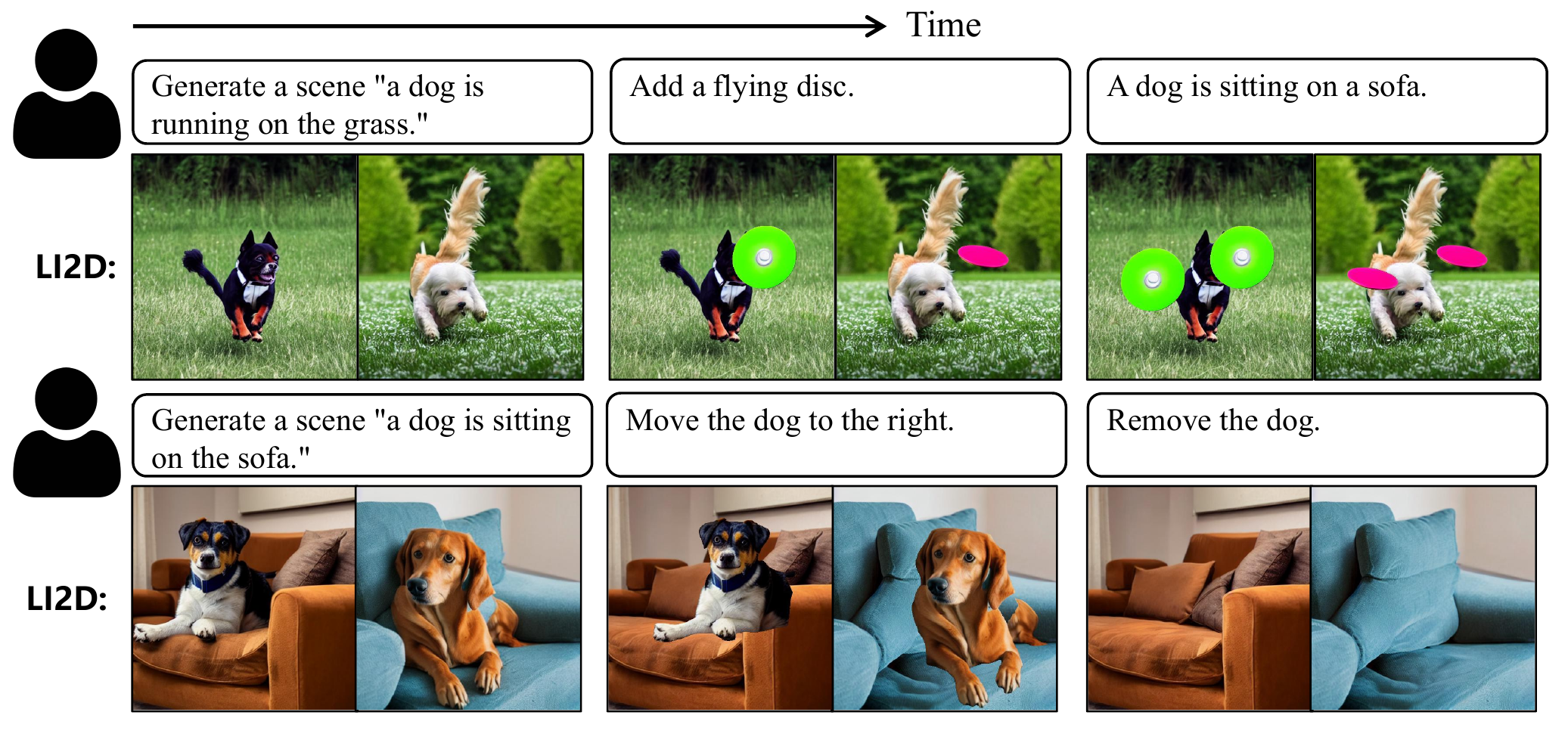}
    \caption{Extending our pipeline into interactive 2D generation and editing.}
    \label{fig:2d}
\end{figure}

Our design pipeline also supports image generation and editing by incorporating Stable Diffusion~\cite{rombach2022high} for layout conditional generation and in-painting, and Segment Anything (SAM)~\cite{kirillov2023segment} for layout element extraction. 
The prompt design is similar to 3D generation; more details are provided in the Appendix. 
Our interactive image generation pipeline is also driven by the layout, and we focus on showcasing the manipulation of elements, including adding, removing, and moving objects defined in the layout.
We use an object described as a unique object identity and define rule-based operations for editing by detecting layout differences.
When adding a new object, we first generate a new image based on the current layout and use SAM to extract the new object, then paste it into the image generated from the previous round interaction. 
When an object is removed, we use SAM to extract the object mask and perform image in-painting. 
When detecting object location or size change, we treat it as object removal and paste masked objects based on the generated layout.
The qualitative results are shown in Fig.~\ref{fig:2d}. The different images are generated using different random seeds. It indicates using LLMs as layout interpreters can be a highly flexible framework for the visual generation.


\section{Related Work}

\noindent \textbf{Interactive Visual Generation and Editing}
Recently, a variety of generation and editing methods~\cite{el2019tell,cui2023i2edit,xia2021tedigan,huang2023ark} have emerged to facilitate multi-round interactions. These methods enable continuous editing of the generative results, ensuring the satisfaction of user requirements throughout the interaction process. 
For instance, TiGAN~\cite{zhou2022tigan} utilizes the pre-trained cross-modal large models CLIP~\cite{radford2021learning} to understand the language feedback from the user to support the successive editing by generative models StyleGAN~\cite{karras2019style}. 
In contrast, Talk-to-Edit~\cite{jiang2021talk} introduces a rule-based system response module to generate responses for multi-round interactions.
Nevertheless, these methods face limitations when it comes to handling non-predefined scenes, which poses a particularly challenging when applying them to complex multi-object 3D scenes.
To tackle this challenge, LLMs offer a promising solution by leveraging their robust capacity to comprehend and reason common knowledge. 
The advancement of LLMs has spurred the development of various approaches to computer vision, which primarily concentrate on tasks related to visual generation~\cite{saharia2022photorealistic,jain2022zero}, and visual understanding~\cite{avrahami2022blended,brooks2022instructpix2pix,haque2023instruct}.
Furthermore, the LLMs are also utilized for interpreting user instructions and translating them into other executable programs.
The VISPROG~\cite{gupta2022visual} leverages LLMs to generate modular programs resembling Python code. These programs are executed to obtain solutions along with clear and interpretable explanations.
In a similar vein, Visual ChatGPT~\cite{wu2023visual}, a system that integrates various Visual Foundation Models, enables users to interact with ChatGPT using images as input and receive images as responses.
Differently, our proposed~\modelname~solves the 3D interactive generation problems by designing a task-specific intermediate representation for LLMs to interpret for better cooperation with generative models.

\noindent \textbf{Text-guided 3D Generative Models}
Recently, there has been growing interest in exploring the use of text-to-3D synthesis~\cite{kim2022datid,poole2022dreamfusion,lin2022magic3d,wang2022score,xu2022dream3d} by leveraging the success of diffusion models in 2D generative models.
Existing methods typically rely on pre-trained text-to-image models~\cite{rombach2022high,saharia2022photorealistic} and employ score distillation sampling to generate 3D geometries and synthesize novel views. 
DreamFusion~\cite{poole2022dreamfusion} exploits the pre-trained text-to-image diffusion model as the prior to update NeRF~\cite{mildenhall2021nerf} for text-to-3D synthesis, which eliminates the need for the 3D training data and demonstrates the effectiveness of this prior in the 3D generation. Similarly, Magic3D~\cite{lin2022magic3d} achieves higher resolution 3D mesh models through a two-stage optimization framework that enhances the text-to-3D synthesis of NeRF. 
Moreover, Latent-NeRF~\cite{metzer2022latent} utilizes score distillation sampling in the latent space to produce 3D results with less computation cost.
Despite these advancements, some challenges~\cite{lin2023componerf} remain including view inconsistency, inaccuracy generation content, and compositional capacity, especially in complex scenes with multi-object generation.

\noindent \textbf{Layout Conditional Visual Synthesis}
The layout-to-image task is an early attempt to perform single-round interactive generation in the image domain.
The layout control in image generation is typically specified as labeled bounding boxes~\cite{zhao2019image,sun2019image,chen2023training,rombach2022high} or semantic maps~\cite{li2023gligen,yang2022reco}.
In this work, we use the labeled bounding boxes as our conditioned input as they can be described through language with specific structure.
While recent advances in large text-to-image models, such as stable diffusion~\cite{rombach2022high}, have made it possible to extend layout-to-image generation task~\cite{chen2023training,li2023gligen} using language.
Despite the progress in the image domain, it remains challenging for the text to convey precise information about the layout of a 3D scene and image generators still have limited spatial fidelity.
To address this issue, there have been several approaches~\cite{cohen2023set,po2023compositional,lin2023componerf} joints employing the conditioning on the text and 3D layout generation from the text. 
These approaches employ different levels of SDS loss to train the local and global scene rendering based on user input layouts.
Our work takes inspiration from ~\cite{lin2023componerf,chen2023training} as they offer distinct advantages in layout conditional generation.
\modelname~eliminates the need for pre-defined object concepts and spatial locations in layout input from the user.
Instead, the natural language input can be automatically interpreted as a layout for the generative models to generate and perform multi-round editing.

\section{Conclusion}
In this paper, we propose a novel interactive language-guided system named \modelname, which integrates LLMs into existing generation models to enable interactive visual content generation in both 2D and 3D domains. By bridging the gap between the natural language and the 3D generative models through LLMs, \modelname~allows users to easily generate and edit visual content, even from continuous language instructions.
To further improve generation quality, we use LLaVA to verify and enrich the generated content based on the rendered views as feedback for LLMs to update the layout.
Our experiments validate the effectiveness of \modelname, demonstrating the potential benefits of incorporating LLMs in generative AI for applications such as metaverse and gaming simulation.
Moreover, our results reveal that LLMs can serve as a visual commonsense knowledge reasoning engine, opening new opportunities for future research.

\noindent \textbf{Broader Impact}
The use of the stable diffusion model in \modelname~brings with it the possibility of inheriting biases or limitations present in the model. This is a concern shared by all generative models, which can potentially be used to create disinformation that is more convincing if presented in 3D. While the synthesized 3D models in \modelname~may not yet match the realism of state-of-the-art 2D image synthesis, they still pose a risk for misuse. Additionally, the automation enabled by generative models such as ours may lead to the displacement of creative workers. However, it's also important to acknowledge the potential for these tools to facilitate growth and enhance accessibility in the creative industry. Overall, it's crucial to consider these ethical implications and use generative models responsibly to maximize their potential benefits while minimizing potential harms.
\bibliographystyle{plain}
\bibliography{arxiv}
\clearpage
\appendix

\section{Implementation Details}
In this section, we provide more model details of the 3D and 2D generative models and more implementation details about integrating our \modelname~and these generative models.

\subsection{Layout Conditional 3D Generative Models}
In \modelname~our 3D generative model adopt the CompoNeRF~\cite{lin2023componerf} which performs multi-object text-to-3D from a configurable layout as input.
In CompoNeRF, the 3D layout requires extracting multiple noun phrases with their corresponding binding attributes and mapping these local text prompts into the corresponding regions.
While in \modelname~we use the LLMs to directly predict such complex 3D layouts from user language input.
The generated layout is defined with multiple local frames associated with a local NeRF as representation.
Each local NeRF contains a local text prompt and a spatial layout described by 3D boxes for each object entity as formulated in our layout format.
In the local NeRF representation, each point $(x_l, y_l, z_l)\in [-1, 1]$ in the local frame is mapped into its corresponding volumetric density $\sigma$ and emitted color $\boldsymbol{C}_l$ using the equation:
$
[\boldsymbol{C}_l, \sigma] = \boldsymbol{\theta}{_l}(x_l, y_l, z_l).
$
Therefore, a local NeRF outputs four pseudo-color channels $\boldsymbol{C}$ and a volume density $\boldsymbol{\sigma}$ when casting a ray through the local frame.
The predicted pseudo-color is then fed into the decoder of Stable Diffusion models to obtain the final rendering result.
During training, CompoNeRF uses Score Distillation Sampling (SDS) proposed in DreamFusion~\cite{poole2022dreamfusion} to propagate the text-to-image generative prior from the diffusion model to the NeRF parameters.
In the original CompoNeRF, a global MLP also is proposed to enhance the global consistency across all the local NeRFs. In our \modelname~, to improve the training efficiency and inheritance of the local NeRF in the generated content, we remove the global MLP for fast convergence. Therefore, replacing certain objects or parts in \modelname~cost less computation. Since the generated layout from the LLMs might be unstable, we perform pre-processing at the box scale, by normalizing all the boxes' boundaries into [-0.8, 0.8].
The weights of the global and local losses are all set to 100 and the learning rates are set to 0.0001.

\subsection{Layout Conditional 2D Generative Models}
In the extension of 2D image generation, our basic models are the open-source Stable Diffusion(v1-5)~\cite{rombach2022high} and Segment Anything~\cite{kirillov2023segment} (SAM-VitH).
In the layout-to-image step, we follow the~\cite{chen2023training} to use backward guidance in the default setting to control the generation process in Stable Diffusion.
During image editing steps with the updated layout as a condition, we first use SAM to segment the objects that detected change in the layout using the center point of the box.
In practice, we use the mask with the second large predicted score to extract the object and dilate the mask with 10 pixels around the mask.
After extracting the object, we perform inpainting using the Stable Diffusion model to obtain the unchanged background based on the updated layout.

\clearpage
\section{Prompt Design for 3D Generation}

\subsection{Scene Generation}
In this section, we provide the context prompt for the scene generations (See Table.~\ref{ct1}) in detail and the case dialog (See Table.~\ref{case1}) for the 3D scene generation task in \modelname.
The key insight is to prompt the LLMs to be specific for the generative models, including the expected sizes of the boxes and the generating of the aligned elements.
Furthermore, we provide the layout visualization (See Figure.~\ref{fig:selayout}) and the more rendering views (See Figure.~\ref{fig:seview}) of the same case across multiple rounds of interaction.
\input{appendix/context3d}

\input{appendix/case1}

\clearpage

\subsection{Single Object Generation}
In this section, we provide the context prompt in detail for the individual objects from multiple compositional parts using language (See Table.~\ref{ct2}) and the case dialog (See Table.~\ref{case2}).
Our main focus is on adjusting or editing specific compositions within individual objects. Moreover, we offer the visualization of the layout for an individual object (See Figure.~\ref{fig:objlayout}), as well as multiple rendering views (See Figure.~\ref{fig:objeview}) of the same case across multiple rounds of interaction. 

\input{appendix/context3dobj}
\input{appendix/case2}

\clearpage
\section{Prompt Design for Generative Feedback}
\begin{figure}[th]
    \centering
    \includegraphics[width=\linewidth]{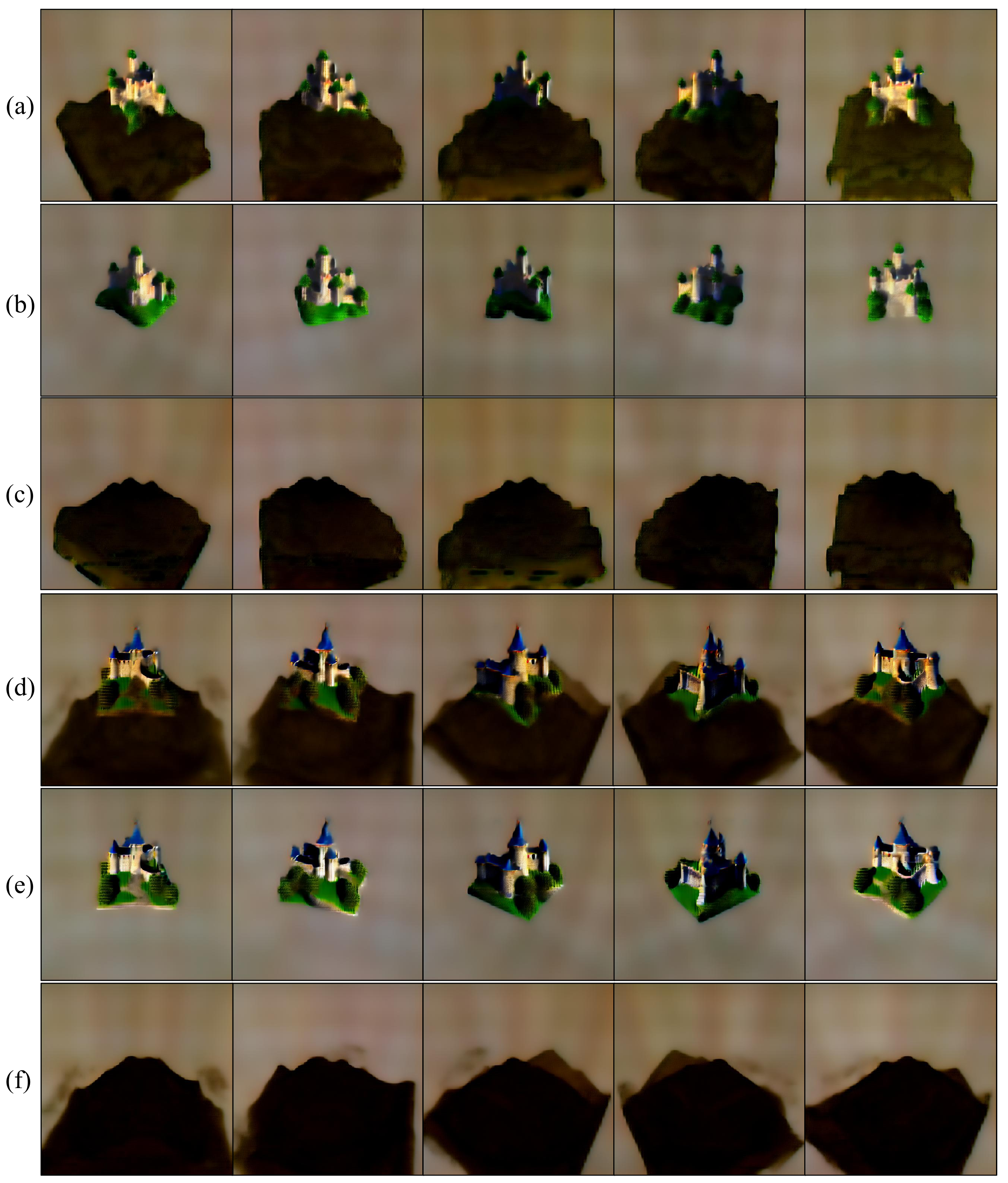}
    \caption{The visualization of the 3D scene generation of generative feedback module. (a) showcases the initial generated scene. (b) and (c) depict the objects present in the initial scene. (d) displays the scene editing performed through feedback. (e) and (f) demonstrate the objects in the updated scene.}
    \label{fig:vis3}
\end{figure}

In this section, we present a comprehensive context prompt for LLaVA to generate feedback on the layout produced by LLMs(See Table.~\ref{ctfb}). The feedback serves as the revised context prompt for LLMs to generate a new layout. 
Furthermore, we provide multiple rendering views (See Figure.~\ref{fig:vis3}) showcasing the original 3D scene and the edited 3D scene.

\input{appendix/case3}
\clearpage
\section{Prompt Design for 2D Generation}
In this section, we showcase the prompt in detail for the LLMs to guide the layout in 2D generation using language (See Table.~\ref{ct2d}) and the case dialog (See Table.~\ref{case4}) for the 2D generation task in \modelname. 

\input{appendix/context2d}
\input{appendix/case4}

\clearpage
\subsection{Prompt Design for Benchmarking Layout Generation}
To evaluate layout generation following the language input qualitatively, we utilize a subset of i-CLEVR~\cite{el2019tell}, a dataset designed for the Generative Neural Visual Artist (GeNeVA) task, as the benchmark.
Each example in i-CLEVR~\cite{el2019tell} dataset consists of a sequence of 5 (image, instruction) pairs. 
Starting from an empty background, each language instruction describes an object to add to the canvas in terms of its shape and color. 
The instruction also describes where the object should be placed relative to existing objects in the scene. 
We sample 50 sequential instructions and their ground-truth arrangement to evaluate the robustness of LLMs following instructions.
As shown in the Table.~\ref{ct5}, we first use the context prompt the same as the 3D generation described in Table.~\ref{ct1}.
Then the LLM is asked to generate the scene layout based on the textual instruction with multi-round sequential instruction (max to 5).
We show a complete cases dialog at Table.~\ref{ct5}. 
Note that in each round of testing, all the instructions are fed simultaneously in a user prompt since we focus on evaluating the layout generation ability based on complex input.
When the instructions are split into multiple rounds of interaction, the capability of maintaining context also needs to be considered making it hard to evaluate the layout generation ability only.
During the numerical evaluation, we calculate the similarity between the front view of the 3D layout and the ground truth images (See Fig.~\ref{fig:iclevrgt}) based on the center points of each object.

\input{appendix/context_inst}
\end{document}

%% file: prompt_design.tex
\refstepcounter{table}
\begin{tcolorbox}[
    standard jigsaw,
    title=Examples of 3D Scene and Single Object Layout Generated by GPT-3.5,
    opacityback=0,
    label=prompt,
    float=t
]
\textbf{(Scene) Context Prompt:} 
... MUST use the following format:

\Context{
\textbf{object\_description}: ['description of single object', 'description of single object'] \\
\textbf{object\_center\_point}: [[0, 0, 0], [0, 0, 0]] \\
\textbf{object\_box\_scales}: [[0.5, 0.5, 0.5], [0.5, 0.5, 0.5]] \\
\textbf{description}: 'description of the whole scene'
}
... (Details are provided in Appendix)

\textbf{(Scene) User Prompt:} 
Generate a scene " a castle on a mountain"

\textbf{(Scene) Generated Response:}

\textcolor{commentcolor}{\textbf{object\_description}: ['a mountain', 'a castle']\\
\textbf{object\_center\_point}: [[0, -0.5, 0], [0, 0, 0]]\\
\textbf{object\_box\_scales}: [[0.9, 0.5, 0.9], [0.4, 0.4, 0.4]]\\
\textbf{description}: 'A castle on a mountain.'}

\textbf{(Object) Context Prompt:} 
...
MUST use the following format:

\Context{
\textbf{part\_description}: [`description of an object part', `description of an object part']\\
\textbf{part\_center\_point}: [[0, 0, 0], [0, 0, 0]]\\
\textbf{part\_box\_scales}: [[0.5, 0.5, 0.5], [0.5, 0.5, 0.5]]\\
\textbf{description}: `description of the whole object'
}
... (Details are provided in Appendix)

\textbf{(Object) User Prompt:} 
Generate an Off-road vehicle

\textbf{(Object) Generated Response:} 

\textcolor{commentcolor}{\textbf{part\_description}: [`vehicle body', `front left wheel', `front right wheel', `rear left wheel', `rear right wheel']\\
\textbf{part\_center\_point}: [[0, 0, 0], [-0.4, -0.25, 0.4], [0.4, -0.25, 0.4], [-0.4, -0.25, -0.4], [0.4, -0.25, -0.4]]\\
\textbf{part\_box\_scales}: [[0.6, 0.3, 1], [0.25, 0.25, 0.25], [0.25, 0.25, 0.25], [0.25, 0.25, 0.25], [0.25, 0.25, 0.25]]\\
\textbf{description}: `an off-road vehicle with a main body and four large wheels'} 
\end{tcolorbox}
\vspace{-10pt}

%% file: appendix/context3d.tex
\refstepcounter{table}
\begin{tcolorbox}[
    standard jigsaw,
    title=Table 4. The Context Prompt of 3D Scene Generation,
    opacityback=0,
    label=ct1,
    float=th
]
\CP

~~~~3DGPT is designed to be able to assist with the 3d modeling tasks, 3DGPT is able to generate a 3d scene based on the input it receives.
When 3DGPT generates a response for a SCENE, MUST use the following format:

```\\
\Context{
object\_description: ['description of single object', 'description of single object']\\
object\_center\_point: [[0, 0, 0], [0, 0, 0]]\\
object\_box\_scales: [[0.5, 0.5, 0.5], [0.5, 0.5, 0.5]]\\
description: 'description of the whole scene'}\\
'''

The number of elements in object\_description, object\_center\_point, and object\_box\_scales should be equal.
Each object\_description should include scene information.
Given an instruction, 3DGPT is able to understand the meaning of the instruction and generate the modification of the object\_description, object\_center\_point or object\_box\_scales based on previous output results.
The relative size of each object should follow the common scene in the world.
The range of object\_center\_point and object\_box\_scales MUST be [-1, 1].
Each object should not have overlap to each other.

\GP

~~~~Understood, how may I assist you today with 3DGPT and 3D modeling tasks?

\end{tcolorbox}

%% file: appendix/case1.tex
\refstepcounter{table}
\begin{tcolorbox}[
    standard jigsaw,
    title=Table 5. A Case Dialog of Multi-Round Interactive 3D Scene Generation,
    breakable,
    opacityback=0,
    label=case1,
    float=htb
]

... (Context Prompt)

\UP

~~~~Generate a scene " a courtyard with a well in the center and several trees"

\GP

~~~~Sure, here's a scene description for a courtyard with a well in the center and several trees:

```\\
\Context{
object\_description: ['a well', 'a tree', 'a tree', 'a tree']\\
object\_center\_point: [[0, 0, 0], [0.7, 0.4, -0.2], [-0.7, 0.3, 0.4], [-0.5, 0.2, -0.6]]\\
object\_box\_scales: [[0.5, 0.5, 0.5], [0.2, 1.5, 0.2], [0.2, 1.5, 0.2], [0.2, 1.5, 0.2]]\\
description: 'A courtyard with a well in the center and several trees'}\\
'''

In this scene, there are four objects: a well in the center, and three trees positioned around it. The position of each object is described by its `object\_center\_point`, which is a list of three values corresponding to the x, y, and z coordinates of the object's center point in the scene. The size of each object is described by its `object\_box\_scales`, which is a list of three values corresponding to the width, height, and depth of the object's bounding box in the scene. Note that the `object\_center\_point` and `object\_box\_scales` values are normalized to the range [-1, 1], and that the size of each object follows the common scene in the world. Also, there is no overlap between any of the objects in the scene.\\

\UP

~~~~Duplicate another well

\GP

~~~~Sure, here's an updated scene description that includes a duplicate well:

```\\
\Context{
object\_description: ['a well', 'a well', 'a tree', 'a tree', 'a tree']\\
object\_center\_point: [[0, 0, 0], [0.3, -0.2, 0.6], [0.7, 0.4, -0.2], [-0.7, 0.3, 0.4], [-0.5, 0.2, -0.6]]\\
object\_box\_scales: [[0.5, 0.5, 0.5], [0.5, 0.5, 0.5], [0.2, 1.5, 0.2], [0.2, 1.5, 0.2], [0.2, 1.5, 0.2]]\\
description: 'A courtyard with two wells in the center and several trees'}\\
'''

In this updated scene, there are five objects: two wells positioned in the center, and three trees positioned around them. The position and size of each object is described using the `object\_center\_point` and `object\_box\_scales` lists, respectively. 

Note that the `object\_description`, `object\_center\_point`, and `object\_box\_scales` lists have been updated to include the second well, and that the `description` field has been updated to reflect the change in the scene.\\

\UP

~~~~Replace all the trees with fir tree

\GP

~~~~Sure, here's an updated scene description that replaces all the trees with fir trees:

```\\
\Context{object\_description: ['a well', 'a well', 'a fir tree', 'a fir tree', 'a fir tree']\\
object\_center\_point: [[0, 0, 0], [0.3, -0.2, 0.6], [0.7, 0.4, -0.2], [-0.7, 0.3, 0.4], [-0.5, 0.2, -0.6]]\\
object\_box\_scales: [[0.5, 0.5, 0.5], [0.5, 0.5, 0.5], [0.2, 1.5, 0.2], [0.2, 1.5, 0.2], [0.2, 1.5, 0.2]]\\
description: 'A courtyard with two wells in the center and several fir trees'}\\
'''

In this updated scene, all three trees have been replaced with fir trees. The position and size of each object remains the same as in the previous scene description, and the `object\_description` and `description` fields have been updated to reflect the change in the scene.

\begin{minipage}[h]{\linewidth}
\includegraphics[width=\linewidth]{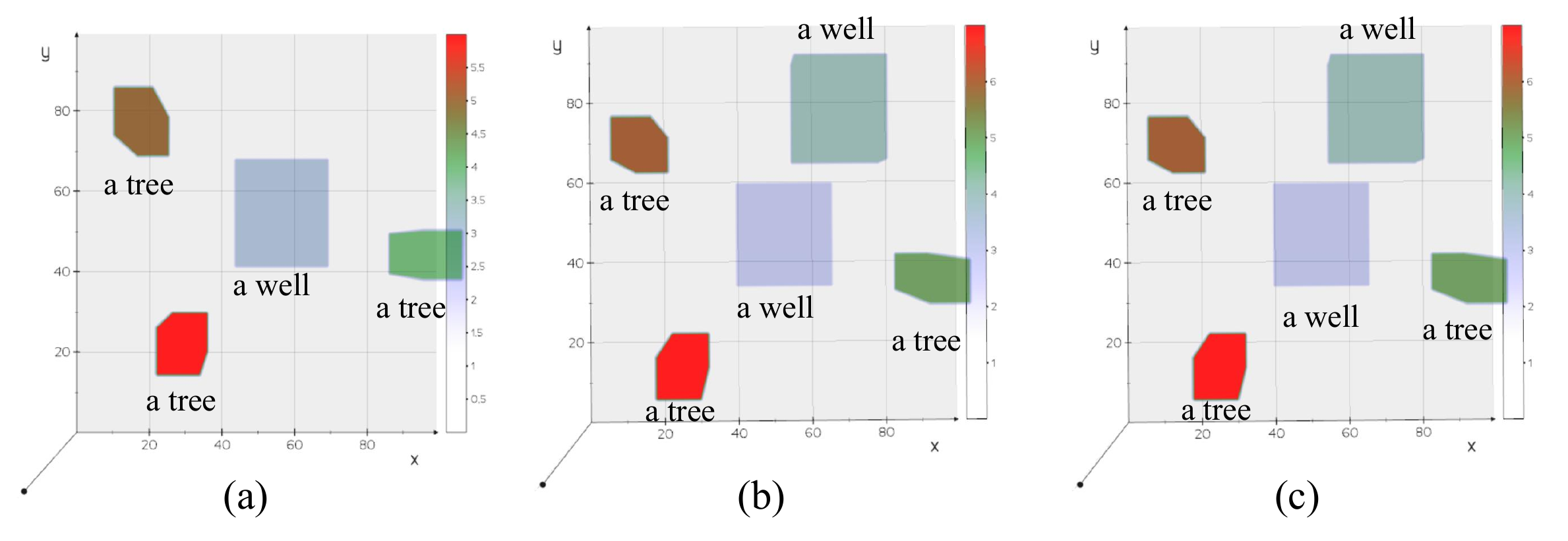}
\captionof{figure}{The visualization of the multiple rounds of generated layouts. (a), (b) and (c) represent the layouts generated for the first, second, and third user prompts. }
\label{fig:selayout}
% (a) illustrates the initial layout generated based on the prompt. (b) demonstrates the addition of a duplicated well to the layout. (c) shows the layout that remains unchanged.

\includegraphics[width=\linewidth]{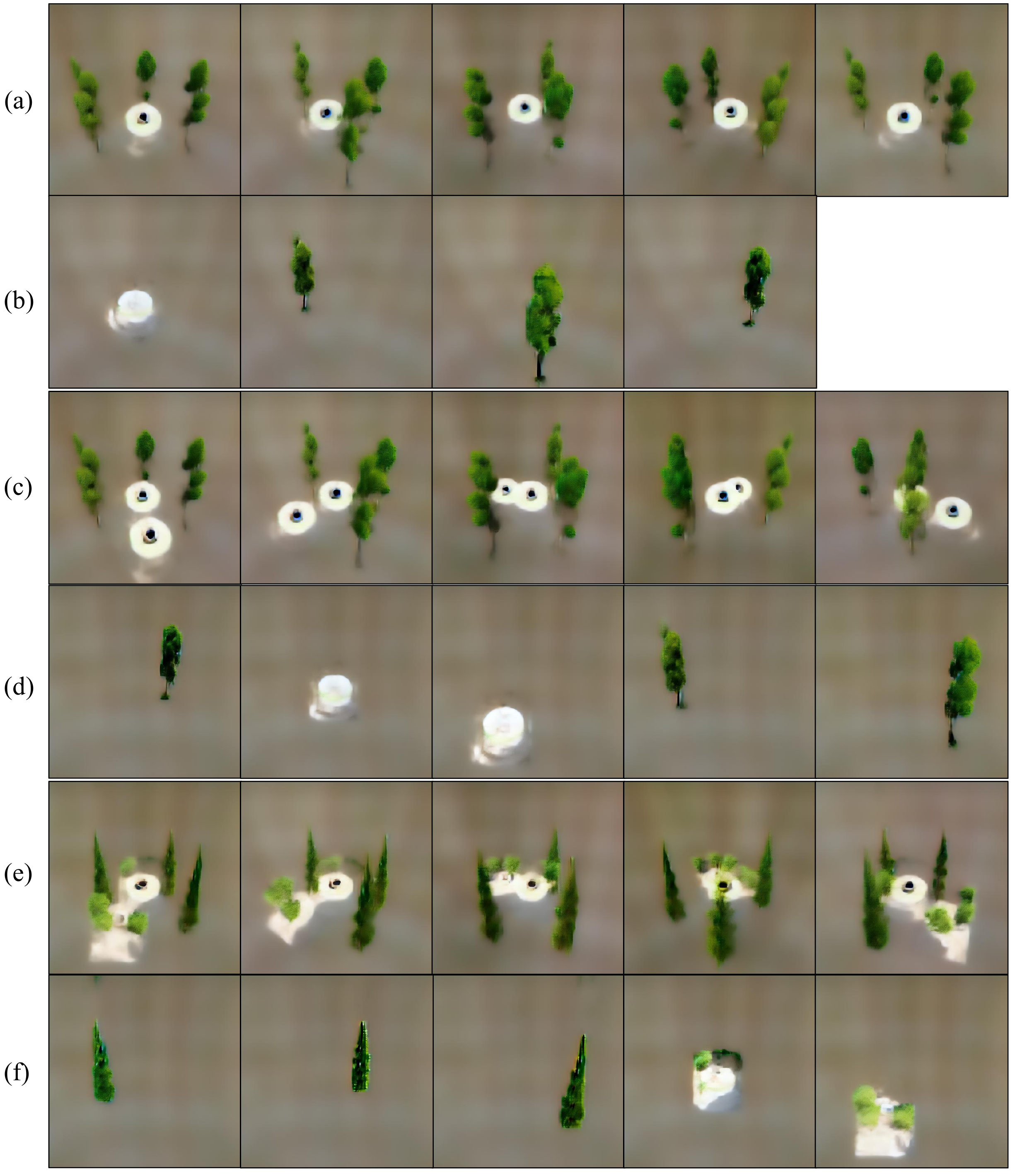}
\captionof{figure}{The visualization of the multiple rounds of generated scenes. (a), (c) and (e) depict the multiple rendering views of the generated scenes from the first, second, and third user prompts. Meanwhile, (b), (d), and (f) showcase the individual generation of objects within these scenes, emphasizing object-level details.}
\label{fig:seview}
\end{minipage}

\end{tcolorbox}

%% file: appendix/context3dobj.tex
\refstepcounter{table}
\begin{tcolorbox}[
    standard jigsaw,
    title=Table 6. The Context Prompt of 3D Single Object Generation,
    opacityback=0,
    label=ct2,breakable,
    float=hb
]
\CP

~~~~3DObjGPT is designed to be able to assist with the 3d modeling tasks, 3DObjGPT is able to generate a 3d scene of object structure based on the input it receives.
When 3DGPT generate a response for a single OBJECT, MUST use the following format:

```\\
\Context{
part\_description: ['description of a object\\
part', 'description of a object part']\\
part\_center\_point: [[0, 0, 0], [0, 0, 0]]\\
part\_box\_scales: [[0.5, 0.5, 0.5], [0.5, 0.5, 0.5]]\\
description: 'description of the whole object'}\\
'''

The number of elements in part\_description, part\_center\_point, and part\_box\_scales MUST be equal.
Each part\_description should include object information.
If parts are duplicated, all of them should be listed.
Given an instruction, 3DObjGPT is able to understand the meaning of the instruction and generate the modification of the object\_description, object\_center\_point or object\_box\_scales based on previous output results.
The relative size of each object should follow the common scene in the world.
The range of object\_center\_point and object\_box\_scales MUST be [-1, 1].

\GP

~~~~Thank you for providing the format for generating responses for a single object using 3DObjGPT. It seems like a very useful tool for 3D modeling tasks.
Based on the given format, the \textbf{part\_description} variable should contain a list of descriptions for each part of the object. Each element in the list should describe a different part of the object, and there should be one element for each part.
The \textbf{part\_center\_point} variable should contain a list of center points for each part of the object. Each element in the list should correspond to the part description in the same position in the \textbf{part\_description} list. The center point represents the position of the center of the part relative to the origin of the object.
The \textbf{part\_box\_scales} variable should contain a list of box scales for each part of the object. Each element in the list should correspond to the part description in the same position in the \textbf{part\_description} list. The box scales represent the size of the bounding box of the part, which can be used to determine the overall size and shape of the object.
The description variable should contain a single description of the whole object, which summarizes the different parts and their relationships to each other.
When generating modifications to the \textbf{object\_description}, \textbf{object\_center\_point}, or \textbf{object\_box\_scales}, the new values should still conform to the same format and rules as the original response. The relative size of each object should still follow the common scene in the world, and the range of \textbf{object\_center\_point} and \textbf{object\_box\_scales} should still be [-1, 1].
\end{tcolorbox}

%% file: appendix/case2.tex
\refstepcounter{table}
\begin{tcolorbox}[
    standard jigsaw,
    title=Table 7. A Case Dialog of Multi-Round Interactive 3D Single Object Generation,
    opacityback=0,breakable,
    label=case2,
    float=hb
]
... (Context Prompt)

\UP

~~~~Generate a windmill

\GP

~~~~Sure, I can generate a windmill for you using 3DObjGPT. Here is an example response:

```\\
\Context{
part\_description: ['base', 'tower', 'blades', 'spinner']\\
part\_center\_point: [[0, -0.5, 0], [0, 0.5, 0], [0, 1.5, 0], [0, 1.5, 0]]\\
part\_box\_scales: [[1, 0.5, 1], [0.2, 1, 0.2], [2, 0.1, 0.5], [0.5, 0.5, 0.5]]\\
description: 'A windmill with a base, tower, blades, and spinner.'}\\
'''

In this response, I have described the windmill as having four parts: the base, tower, blades, and spinner. The `part\_center\_point` values describe the position of each part relative to the origin of the windmill. The `part\_box\_scales` values describe the size and shape of each part. 

The base of the windmill is a rectangular prism with dimensions of 1 unit in the x and z directions and 0.5 units in the y direction. The tower is a thin cylinder with a height of 1 unit and a radius of 0.1 units. The blades are a long, flat rectangular prism with dimensions of 2 units in the x direction, 0.1 units in the y direction, and 0.5 units in the z direction. The spinner is a small cube with dimensions of 0.5 units in each direction.

Of course, this is just one possible representation of a windmill. Depending on the level of detail and style that you want, there are many other ways that a windmill could be described using 3DObjGPT.

\UP

~~~~Make the tower more wider

\GP

~~~~Sure, I can modify the windmill tower to make it wider. Here is an updated response:

```\\
\Context{
part\_description: ['base', 'tower', 'blades', 'spinner']\\
part\_center\_point: [[0, -0.5, 0], [0, 0.5, 0], [0, 1.5, 0], [0, 1.5, 0]]\\
part\_box\_scales: [[1, 0.5, 1], [0.4, 1, 0.4], [2, 0.1, 0.5], [0.5, 0.5, 0.5]]\\
description: 'A windmill with a base, tower, blades, and spinner.'}\\
'''

In this modified response, I have updated the `part\_box\_scales` value for the tower to be `[0.4, 1, 0.4]`, which means the tower has a wider radius of 0.4 units instead of 0.1 units in the original response. The `part\_description` and `part\_center\_point` values remain the same.

Note that I have kept the other parts of the windmill the same in this modified response. If you want me to modify any other aspects of the windmill, feel free to let me know!

\UP

~~~~Duplicate the blades in the middle of tower

\GP

~~~~Sure, I can duplicate the blades in the middle of the tower for you. Here is an updated response:

```\\
\Context{
part\_description: ['base', 'tower', 'blades', 'blades', 'spinner']\\
part\_center\_point: [[0, -0.5, 0], [0, 0.5, 0], [0, 1.5, 0.5], [0, 1.5, -0.5], [0, 1.5, 0]]\\
part\_box\_scales: [[1, 0.5, 1], [0.4, 1, 0.4], [2, 0.1, 0.5], [2, 0.1, 0.5], [0.5, 0.5, 0.5]]\\
description: 'A windmill with a base, tower, blades, blades, and spinner.'}\\
'''

In this modified response, I have added a second `blades` part to the `part\_description` list, and added a corresponding `part\_center\_point` value of `[0, 1.5, -0.5]` to represent the position of the duplicated blades. I have also updated the `part\_box\_scales` values for both sets of blades to be `[2, 0.1, 0.5]`, which means that both sets of blades are now the same size.

Note that I have kept the other parts of the windmill the same in this modified response. If you want me to modify any other aspects of the windmill, feel free to let me know!

\begin{minipage}[h]{\linewidth}
\includegraphics[width=0.9\linewidth]{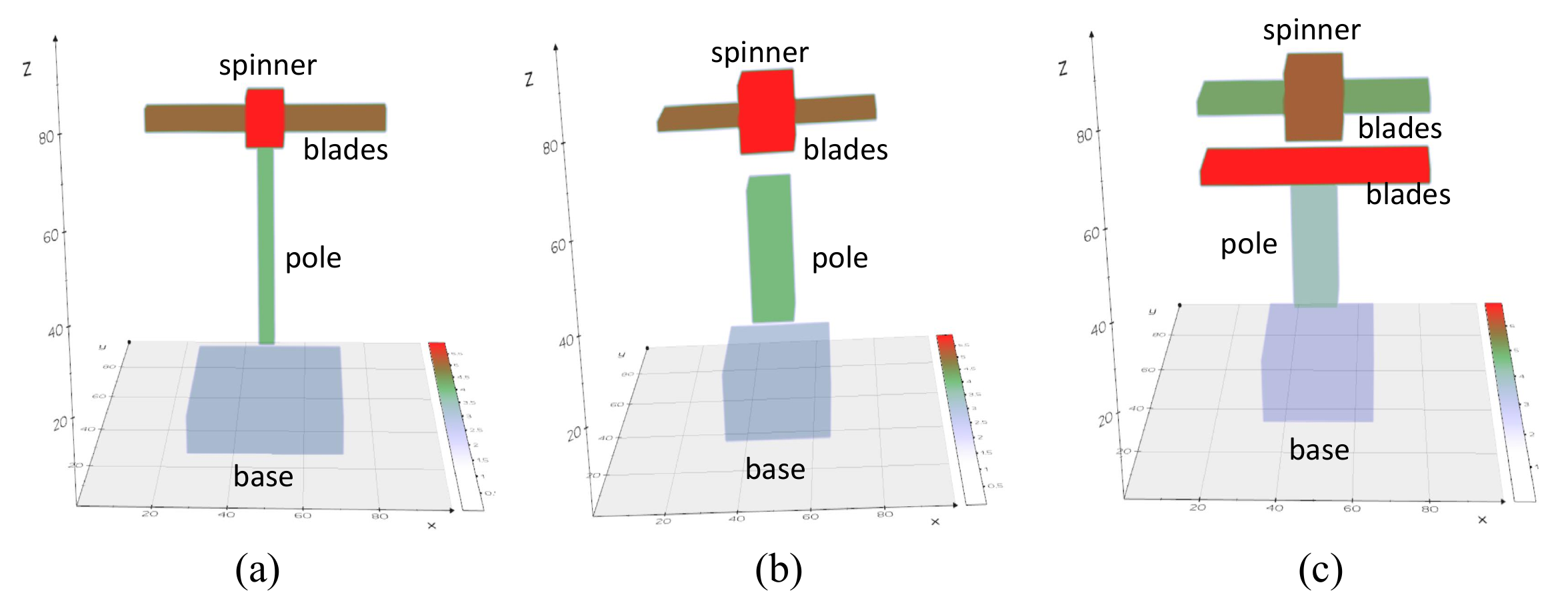}
\captionof{figure}{The visualization of the multiple rounds of a single object from compositional parts guided by the prompt. (a), (b), and (c) represent the compositional parts of the object generated for the first, second, and third user prompts.}
\label{fig:objlayout}

\includegraphics[width=\linewidth]{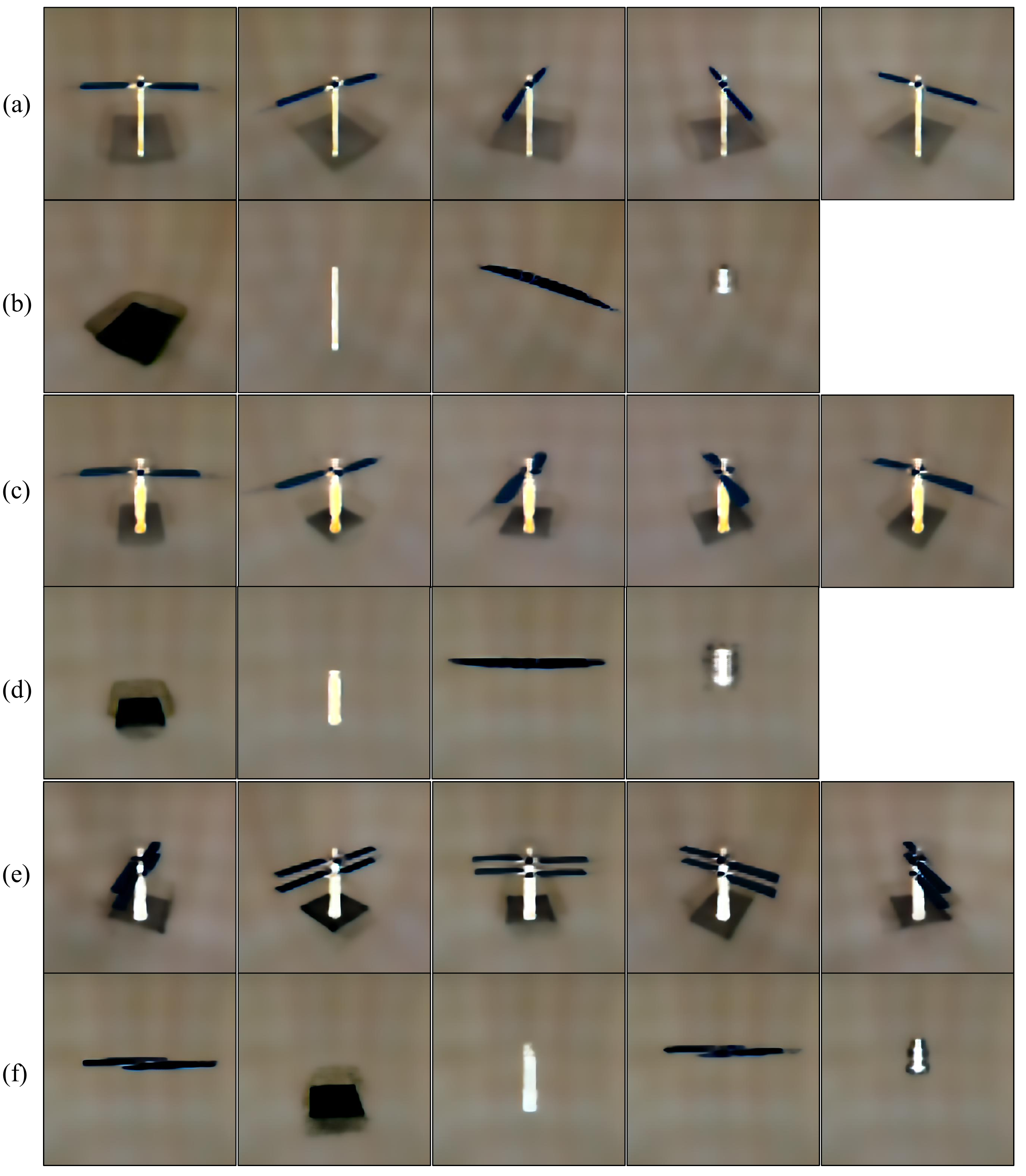}
\captionof{figure}{The visualization of the multiple rounds of 3D single object generation. (a), (c), and (e) depict multiple rendering views of the generated single object from the first, second, and third user prompts. Furthermore, (b), (d), and (f) showcase the individual generation of compositional parts that form the object, highlighting component-level details.}
\label{fig:objeview}
\end{minipage}
\end{tcolorbox}

%% file: appendix/case3.tex
\refstepcounter{table}
\begin{tcolorbox}[
    standard jigsaw,
    title=Table 8. The Context Prompt and Case Dialog of 3D Scene Generation,
    opacityback=0,
    label=ctfb, breakable,
    float=hb
]
\textbf{Generated Layout:}

\Context{
object\_description: ['a mountain', 'a castle']\\
object\_center\_point: [[0, -0.5, 0], [0, 0, 0]]\\
object\_box\_scales: [[0.9, 0.5, 0.9], [0.4, 0.4, 0.4]]\\
description: 'A castle on a mountain.'
}\\

\CP(LLaVA)

~~~~Does the image match the description \yellow{\$\{DESCRIPTION\}~("A castle on a mountain.")}

\LP

~~~~The image does not match the description "A castle on a mountain." Instead, it shows a computer-generated (CGI) image of a castle with a grassy hill. There is no indication of the castle being situated on a mountain in the provided image.

\CP(LLaVA)

~~~~Generate the description with more details to match the scene \yellow{\$\{DESCRIPTION\}~("A castle on a mountain.")}

\LP
(FEEDBACK)

~~~~To modify the description to match the "A castle on a mountain," you could add specific details about the mountainous environment surrounding the castle. For example, you could include sentences such as "The castle is situated atop a tall mountain, overlooking the surrounding landscape," or "The castle is built into the side of a massive mountain, providing a sense of natural fortification." By incorporating these details, you can create a more accurate and visually appealing image that aligns with the original description.

\CP(LLMs)

~~~~Given feedback \yellow{\$\{FEEDBACK\} ("To modify the description to match the "A castle on a mountain," you could add specific details about the mountainous environment surrounding the castle. For example, you could include sentences such as "The castle is situated atop a tall mountain, overlooking the surrounding landscape," or "The castle is built into the side of a massive mountain, providing a sense of natural fortification." By incorporating these details, you can create a more accurate and visually appealing image that aligns with the original description.")} update the layout.

\GP~(LLMs)

~~~~Sure, here's an updated layout that includes the specific details about the mountainous environment surrounding the castle:

```\\
\Context{
object\_description: ['a mountain', 'a castle']\\
object\_center\_point: [[0, -0.5, 0], [0, 0, 0]]\\
object\_box\_scales: [[0.9, 0.5, 0.9], [0.4, 0.4, 0.4]]\\
description: 'The castle is situated atop a tall mountain, overlooking the surrounding landscape. The castle itself is built into the side of the mountain, providing a sense of natural fortification. A winding path leads up to the castle gates, which are guarded by two towering statues. From the castle walls, one can see for miles in every direction, taking in the majesty of the surrounding peaks and valleys.'}\\
'''

In this updated layout, the scene is described in more detail. The castle is situated on a tall mountain, and is built into the side of the mountain for natural fortification. The castle gates are guarded by two towering statues, and a winding path leads up to the gates. From the castle walls, one can see for miles in every direction, taking in the majesty of the surrounding peaks and valleys.

\end{tcolorbox}

%% file: appendix/context2d.tex
\refstepcounter{table}
\begin{tcolorbox}[
    standard jigsaw,
    title=Table 9. The Context Prompt of 2D Image Generation,
    opacityback=0,
    label=ct2d,breakable,
    float=hb
]
\CP 
2DGPT is designed to be able to assist with the 2d modeling tasks, 2DGPT is able to generate a 2d object layout using bounding box based on the input it receives, please use the following format:

```\\
\Context{
object\_description: ['a object description', 'a object description']\\
object\_center\_point: [[0.1, 0.2], [0.3, 0.4]]\\
object\_scale: [[0.9, 0.9], [0.7, 0.8]]\\
description: 'description of the whole scene'}\\
'''\\
Each object\_center\_point and object\_scale are normalized and MUST be [0, 1]. The number of elements in object\_description and object\_box should be equal. When input a instruction, 2DGPT is able to understand the meaning of instruction and reason the modification of the locations or size or description by output the same format as input. The relative size of each object should follow the common scene in the world and the object usually are close to the camera. Each width and height of object MUST large than 0.2. In the scene, the bottom part usually is the ground.

\GP
Sure, I can assist with 2D modeling tasks using the given format. Please provide me with more details on the task you want me to perform.

\end{tcolorbox}

%% file: appendix/case4.tex
\refstepcounter{table}
\begin{tcolorbox}[
    standard jigsaw,
    title=Table 10. A Case Dialog of Multi-Round Interactive 2D Image Generation,
    breakable,
    opacityback=0,
    label=case4,
    float=hb
]

... (Context Prompt)

\UP

~~~~Generate a scene "a dog is sitting on the sofa"

\GP

~~~~Based on the input you provided, I can generate a 2D object layout using the following format:

```\\
\Context{
object\_description: ['dog', 'sofa']\\
object\_center\_point: [[0.5, 0.5], [0.5, 0.75]]\\
object\_scale: [[0.6, 0.6], [0.8, 0.2]]\\
description: 'a dog is sitting on the sofa'}\\
'''

The above structure represents the following:
- There are two objects in the scene with descriptions 'dog' and 'sofa' respectively.
- The center point of the dog object is [x=0.5, y=0.5], which represents its position on a 2D plane where (0,0) is the top left corner and (1,1) is the bottom right corner of the image.
- The scale of the dog object is [width=0.6, height=0.6], which represents the size of the bounding box around the object.
- The center point of the sofa object is [x=0.5, y=0.75], which represents its position on the same 2D plane.
- The scale of the sofa object is [width=0.8, height=0.2], which represents the size of the bounding box around the object.
- The scene description is 'a dog is sitting on the sofa'.
Note that the bottom part of the scene is assumed to be the ground. The relative size of the objects follows the common scene in the world, with the sofa being relatively larger than the dog. Also, the dog's bounding box is slightly larger than the minimum required width and height of 0.2, to ensure that the object is clearly visible in the scene. Let me know if you would like me to modify the location, size, or description of any object in the scene.

\UP

~~~~Move the dog to the right 

\GP

~~~~Sure, I can modify the 2D object layout as per your request. Here's the updated object layout with the dog moved to the right:

```\\
\Context{
object\_description: ['dog', 'sofa']\\
object\_center\_point: [[0.7, 0.5], [0.5, 0.75]]\\
object\_scale: [[0.6, 0.6], [0.8, 0.2]]\\
description: 'a dog is sitting on the sofa'}\\
'''

In the above layout, the center point of the dog object has been modified to [x=0.7, y=0.5], which represents its new position to the right of its original position. The rest of the layout remains the same.

\UP

~~~~Remove the dog

\GP

~~~~Sure, I can modify the 2D object layout as per your request. Here's the updated object layout with the dog removed:

```\\
\Context{
object\_description: ['sofa']\\
object\_center\_point: [[0.5, 0.75]]\\
object\_scale: [[0.8, 0.2]]\\
description: 'there is a sofa'}\\
'''

In the above layout, the dog object has been removed, and the object descriptions, center points, and scales have been adjusted accordingly. The scene description has also been modified to reflect the absence of the dog.

\end{tcolorbox}

%% file: appendix/context_inst.tex
\refstepcounter{table}
\begin{tcolorbox}[
    standard jigsaw,
    title=Table 11. The Prompt Design and Cases Dialog of Benchmarking on i-CLEVR,
    opacityback=0,
    label=ct5, breakable,
    float=hb
]
\CP ...(Please refer to Table.~\ref{ct1})

\GP

Understood. How may I assist you with your 3D modeling tasks?

\UP (1-Round)

Add a cyan cube at the center

\GP

```\\
\Context{
object\_description: ['Cyan cube']\\
object\_center\_point: [[0, 0, 0]]\\
object\_box\_scales: [[0.5, 0.5, 0.5]]\\
description: 'A cyan cube at the center.'
}
'''

Let me know if you have any further modifications or adjustments.

\CP ...(Please refer to Table.~\ref{ct1})

\UP (2-Round)

Add a cyan cube at the center\\
Add a red cylinder behind it\\

\GP

```\\
\Context{
object\_description: ['Cyan cube', 'Red cylinder']\\
object\_center\_point: [[0, 0, 0], [0, 0, -0.8]]\\
object\_box\_scales: [[0.5, 0.5, 0.5], [0.3, 0.3, 0.6]]\\
description: 'A cyan cube at the center and a red cylinder behind it.'
}
'''

Let me know if you have any further modifications or adjustments.
\allowbreak
\CP ...(Please refer to Table.~\ref{ct1})

\UP (3-Round)

Add a cyan cube at the center\\
Add a red cylinder behind it\\
Add a cyan sphere in front of it on the left and in front of the cyan cube on the left

\GP

```\\
\Context{
object\_description: ['Cyan cube', 'Red cylinder', 'Cyan sphere']\\
object\_center\_point: [[0, 0, 0], [0, 0, -0.8], [-0.6, 0, 0.4]]\\
object\_box\_scales: [[0.5, 0.5, 0.5], [0.3, 0.3, 0.6], [0.3, 0.3, 0.3]]\\
description: 'A cyan cube at the center, a red cylinder behind it, and a cyan sphere in front of it on the left and in front of the cyan cube on the left.'
}
'''

Let me know if you have any further modifications or adjustments.

\CP ...(Please refer to Table.~\ref{ct1})

\UP (4-Round)

Add a cyan cube at the center\\
Add a red cylinder behind it\\
Add a cyan sphere in front of it on the left and in front of the cyan cube on the left\\
Add a purple cylinder behind it on the left and in front of the cyan cube on the left

\GP

```\\
\Context{
object\_description: ['Cyan cube', 'Red cylinder', 'Cyan sphere', 'Purple cylinder']\\
object\_center\_point: [[0, 0, 0], [0, 0, -0.8], [-0.6, 0, 0.4], [-0.6, 0, -0.4]]\\
object\_box\_scales: [[0.5, 0.5, 0.5], [0.3, 0.3, 0.6], [0.3, 0.3, 0.3], [0.3, 0.3, 0.6]]\\
description: 'A cyan cube at the center, a red cylinder behind it, a cyan sphere in front of it on the left and in front of the cyan cube on the left, and a purple cylinder behind it on the left and in front of the cyan cube on the left.'
}
'''

Let me know if you have any further modifications or adjustments.

\CP ...(Please refer to Table.~\ref{ct1})

\UP (5-Round)

Add a cyan cube at the center\\
Add a red cylinder behind it\\
Add a cyan sphere in front of it on the left and in front of the cyan cube on the left\\
Add a purple cylinder behind it on the left and in front of the cyan cube on the left\\
Add a red cube behind it on the right and in front of the red cylinder on the right

\GP

```\\
\Context{
object\_description: ['Cyan cube', 'Red cylinder', 'Cyan sphere', 'Purple cylinder', 'Red cube']\\
object\_center\_point: [[0, 0, 0], [0, 0, -0.8], [-0.6, 0, 0.4], [-0.6, 0, -0.4], [0.6, 0, 0]]\\
object\_box\_scales: [[0.5, 0.5, 0.5], [0.3, 0.3, 0.6], [0.3, 0.3, 0.3], [0.3, 0.3, 0.6], [0.5, 0.5, 0.5]]\\
description: 'A cyan cube at the center, a red cylinder behind it, a cyan sphere in front of it on the left and in front of the cyan cube on the left, a purple cylinder behind it on the left and in front of the cyan cube on the left, and a red cube behind it on the right and in front of the red cylinder on the right.'
}
'''

Please let me know if you need any modifications or adjustments.

\includegraphics[width=\linewidth]{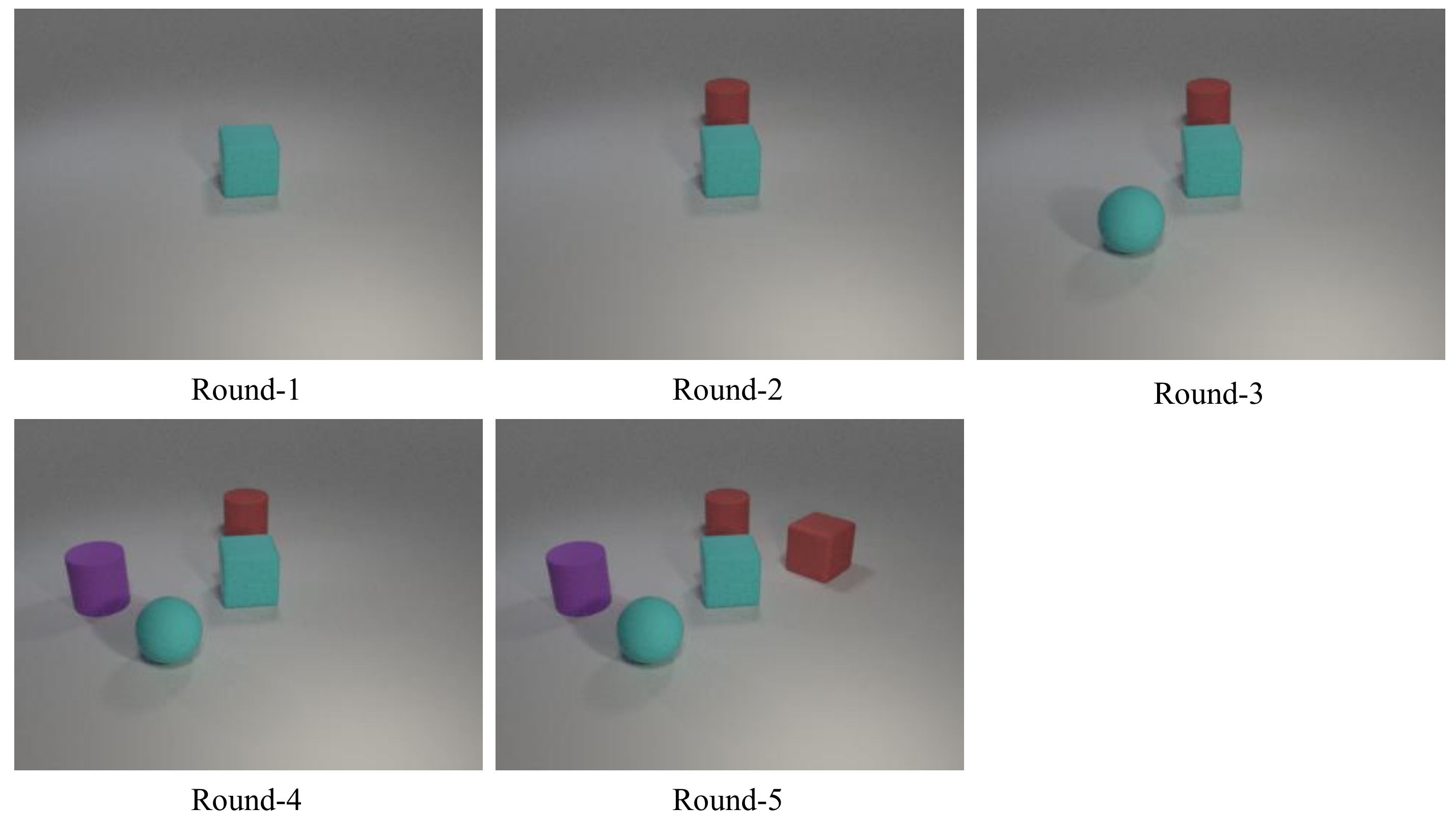}
\captionof{figure}{The visualization of the ground truth image of the multiple rounds instruction of i-CLEVR.}
\label{fig:iclevrgt}

\end{tcolorbox}